# Features Based Adaptive Augmentation for Graph Contrastive Learning

**Adnan Ali**
University of Science and Technology of China
Hefei, Anhui, China, 230006
adnanali@mail.ustc.edu.cn

**Jinlong Li**
University of Science and Technology of China
Hefei, Anhui, China, 230006
jlli@ustc.edu.cn

## Abstract

Self-Supervised learning aims to eliminate the need for expensive annotation in graph representation learning, where graph contrastive learning (GCL) is trained with the self-supervision signals containing *data-data* pairs. These data-data pairs are generated with augmentation employing stochastic functions on the original graph. We argue that some features can be more critical than others depending on the downstream task, and applying stochastic function uniformly, will vandalize the influential features, leading to diminished accuracy. To fix this issue, we introduce a Feature Based Adaptive Augmentation (FebAA) approach, which identifies and preserves potentially influential features and corrupts the remaining ones. We implement FebAA as plug and play layer and use it with state-of-the-art Deep Graph Contrastive Learning (GRACE) and Bootstrapped Graph Latents (BGRL). We successfully improved the accuracy of GRACE and BGRL on eight graph representation learning's benchmark datasets.

## 1 Introduction

Graphs are connected yet complicated data structures, and to apply machine learning algorithms like classification and clustering, graphs are embedded in euclidean space while preserving their connectivity and attributive information. Graph Representation Learning (GRL) aims to embed the graphs where Graph Neural Networks (GNN) [17] emerged as significant contributors against the matrix factorization-based methods and random-walk based algorithms [3]. Most of the available GRL-GNN research is based on semi-supervised learning, with a bottleneck of time and cost expensive data annotation. A practical solution to this bottleneck became prominent in the deep learning community, namely Self-Supervised Learning (SSL) [18]. SSL methods can be divided into two categories based on their working principle: contrastive and predictive models, trained with data-data and data-label pairs, respectively [18]. Graph Contrastive Learning (GCL) emerged from contrastive learning, is the focus of this paper.

The GCL paradigm can be divided into four dimensions [23], Data Augmentation, Contrastive Modes, Contrastive Objective, and Negative Mining Strategies. As the first of the four dimensions, data augmentation is an essential but undermined dimension. Data augmentation is to create views where the graph should preserve its important and downstream-task relevant attributes and structure [19, 25]. To generate multiple views from the original graph, data augmentation functions, e.g., node insertion [20], dropping [19], edge insertion[14, 4, 20], deletion [14, 20, 25], perturbation [14, 19], diffusion [4, 21, 8], ego-nets [5, 22, 1], random walks [19, 12, 4], importance sampling[6] are performed on *Topology Level* and attribute masking [24, 4, 13], and shuffling[14, 10, 7] are applied on *Features Level*. The empirical study [23] yields that the performance of GCL is highly dependent on the choice of topology augmentation functions, and previous work [14, 24, 4, 25, 13] sheds light on data augmentation importance by describing the graph corruption as an open problem needing attention. Although there are multiple augmentation functions, most of studies prefer to apply these functions in stochastic manners for simplicity and to reduce processing time, resulting in the undermined data augmentation dimension. We investigate and improve the data augmentation with GRACE [24] and BGRL [13] because of their different working principle from each other, dual branch augmentation, and the better accuracy in published work.

As the state-of-the-art models, both GRACE [24] and BGRL [13] have different working principles but use the same type of augmentation functions to corrupt the graph, i.e., for feature level augmentation, random features are masked with zeros, and edges are dropped randomly for topology augmentation. BGRL and GRACE's prime focus is to propose a graph representation learning framework where data augmentation is performed as an essential requirement. GRACE only use stochastic augmentation functions, while for further experiments, BGRL also uses adaptive augmentation presented in GCA [25]; however, describes it as "complex and not useful". We argue that if GCA's adaptive augmentation is not much useful for BGRL, we should not exclude the possibility of improvement with other adaptive augmentation methods. For example, our proposed method evaluates adaptive augmentation as "useful" on both BGRL and GRACE on eight graph contrastive learning benchmark datasets.

We argue that some data features can be more influential than others for different downstream tasks, and masking less influential features while preserving essential features will improve accuracy compared to uniform masking. Motivated by variable importance analysis (VIA) techniques, stating that "input variables (features) have an obvious effect on output variables"[15] and GNN Explainers [9] working principle, which finds the influential elements of the graph. Based on GNN Explainers and VIA, we present the *Features Based Adaptive Augmentation* (FebAA) to identify the influential features as their influence on the downstream task at first. Then, the features are divided into two sections, *Preserved Features* are features that remain untouched, and *Candidate Features* are features candidates for masking. At *Last*, the masking is performed from the candidate features. Implementation, experiment codes and hyper-parameters of FebAA can be found at https://github.com/mhadnanali/FebAA

Contributions of this work are:

1. We propose Features Based Adaptive Augmentation (FebAA) in graph contrastive learning paradigm, where feature influence is computed per downstream task with an algorithm; then, we split the features into preserved and candidate features and create the views by masking from candidate features. Furthermore, we provide the list of features as per their importance for the benchmark datasets.

2. We introduced adaptive augmentation as a plug-and-play layer that can be used with any GCL method on any feature based dataset. By incorporating into the state of the art methods: GRACE [24] and BGRL [13], our FebAA improved 4.22% to 5.12% F1 score on Cora and CiteSeer datasets with GRACE and 0.20% to 1.02% on five benchmark datasets with BGRL.

## 2 Preliminaries

Let $G = (V, E)$ be a graph where $V = \{v_1, v_2, v_3, \ldots, \}$ is node set and $E \subseteq (V \times V)$ is edge set. The attributed graph refers to a graph, in which nodes or edges have their own attributes a.k.a features [16] and represented as $Q = (X, A)$, where $X \in \mathbb{R}^{N \times F}$ is node feature matrix and $A \in \{0, 1\}^{N \times N}$ is adjacency matrix, $N = |V|$ is the number of nodes of $G$ and $F$ represents number of features. Furthermore, $x_{ij} \in X$, $i = \{1, 2, \ldots, N\}$ and $j = \{1, 2, \ldots, F\}$ denotes value of the $j$-th feature of node $v_i$ and $\boldsymbol{x}_j$ are the $j$-th columns of $X$. The learned representation is presented as $H = \{h_1, h_2, \ldots, \}^T = f(X, A) \in \mathbb{R}^{N \times F'}$ where $h_i$ is the embedding of node $v_i$ and $f(\cdot)$ is the encoder to convert graph into low dimensional space.

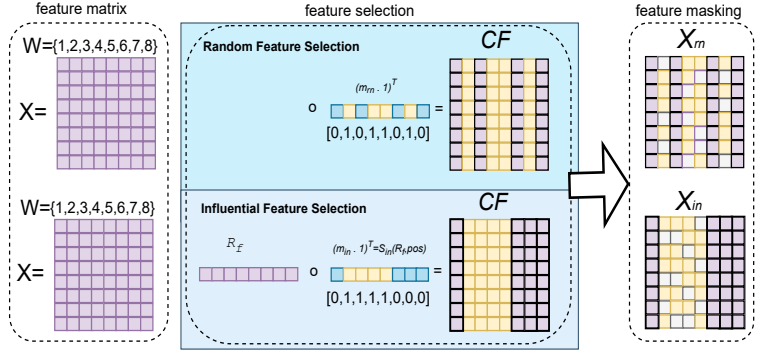

Figure 1: Feature selection have two modules; random feature selection: features are selected randomly from $W$, while in influential feature selection, an algorithm is used to find the $R_f$. Based on $m_{rn}$ or $m_{in}$, candidate feature $CF$ (yellow columns) is created, which is used for feature masking

## 3 Feature based adaptive augmentation (FebAA)

Some features can be more critical than others for a specific downstream task in a graph. To identify and preserve the important features and perform augmentation functions on less important ones, we introduce the Feature Based Adaptive Augmentation (FebAA). FebAA parts features into two subsets *preserved* and *candidate* features, and then performs masking from the candidate features.

As shown in figure 1, FebAA has two modules: *feature selection* and *feature masking*. Former identifies the subset named *Candidate Features* denoted as $CF$ from feature matrix $X$, and the latter performs masking on the $CF$.

### 3.1 Feature selection

Feature selection is to choose the subset of *candidate features*, as illustrated in Figure 1. Feature selection is performed only once, while the next section, feature masking is performed at every training epoch. Let all node features be denoted by $W = \{1, 2, 3, \ldots, F\}$ and its subset $CF \subset W$ is defined as candidate features, determined by Equation 1 and illustrated in Figure 1 in yellow color.

$$CF = W \circ (m_{rn} \oplus m_{in}) \quad (1)$$

where $\circ$ is element-wise multiplication, $\oplus$ indicates that between $m_{rn}$ and $m_{in}$ either one can be true but not both, $m_{rn} \in \{0, 1\}^F$ and $m_{in} \in \{0, 1\}^F$ are indicator vectors whose element value '1' indicates a feature is selected and '0' indicates as not selected, and the number of '1's depends on $F_r^s$. Feature masking ratio $F_r^s$ is a hyper-parameter value denoted as $F_r^s = \{1\%, 2\%, \ldots, 99\%\}$, it decides the candidate feature subset size. The value for $m_{rn}$ is extracted randomly from uniform distribution, and value of $m_{in}$ is decided by influence.

**Influential feature selection:** Influential feature selection is an adaptive behavior that chooses the features as per their influence on downstream tasks. Algorithm 1 finds the influential features and returns a features ranking set $R_f$ containing sorted features based on their influence on the downstream task.

Algorithm 1 takes five steps to find $R_f$; *Step-1* takes $j$-th feature of $X$, replaces it with zeros, and stores it as a temporary feature matrix $(\bar{T})$. *Step-2* train the model on $(\bar{T})$ using any *selected machine learning model* and save the results in a temporary array. *Step-3* is to repeat the first two steps the $j$ times. *Step-4* is to repeat all three steps $n$ times and take an average. The last *step* is to sort the averaged results into ascending order and store them in the features ranking set $Rf$. In *step-2* the *selected machine learning model* we chose is BGRL with a low setting (150 epochs), while it is an open research area to find influential features, and we leave it to readers if they want to choose alternative method.

Furthermore, $R_f$ is used in Equation 2, which illustrates the influential feature selection process.

$$\boldsymbol{m}_{in} = S_{in}(R_f, pos) \quad (2)$$

where $S_{in}(\cdot)$ is a function that takes $R_f$ and $pos$ then returns the $m_{in}$ illustrated in Figure 1, $R_f$ is the set of influential features, and $pos$ indicates the starting position of $R_f$. *Starting position pos* is a hyper-parameter value denoted as $pos = \{L, M\}$ decides whether candidate features should be selected from least output resulting $L$ features or most output resulting $M$ features from the $R_f$.

To exemplify the working of the feature selection section: First, for *random feature selection*, let $|W| = 10$ and $F_r^s = 60\%$, which indicates that 6 out of 10 features will be *randomly* selected as candidate features. Second, for *influential feature selection* starting position is also required; let $pos = L$, $|W| = 10$ and $F_r^s = 60\%$, which indicates that 6 out of features with the least influence will be selected as candidate features. In both cases, the cardinality of $CF$ will be $|CF| = 6$.

---

**Algorithm 1** Feature ranking algorithm

---

1: $n \leftarrow 1$ ▷ Number of rounds
2: $j \leftarrow 0$ ▷ To loop through features
3: $tempResults \leftarrow [\,][\,]$ ▷ Temp variable
4: $x \leftarrow (|X| - 1)$
5: **while** $n \geq 3$ **do**
6:    **for** $j \geq x$ **do**:
7:       $\overline{T} \leftarrow$ Mask($j^{th}$ Feature) ▷ Temp feature matrix $\overline{T}$
8:       $results[n][j] \leftarrow ApplyMethod(\overline{T})$ ▷ Apply any suitable ML method
9:    **end for**
10: **end while**
11: $AResults \leftarrow Average(tempResults[n])$. ▷ Average results based on rounds $n$
12: $R_f \leftarrow Sort(AResults)$ ▷ Sort in ascending order

---

### 3.2 Feature masking

Feature masking is a process to mask features from the candidate features $CF$; we follow GCA [25] like approach for masking by adding noise to node attributes via randomly masking a fraction of dimensions with zeros in candidate features. Features are masked from $CF$, while how many features will be masked depends on Bernoulli distribution $B(1 - F_p^s)$ followed by feature masking probability $F_p^s$ value. *Feature masking probability* is a hyper-parameter value denoted as $F_p^s = \{1\%, 2\%, \ldots, 100\%\}$, which decides the masking probability to mask the features from candidate features.

### 3.3 Plug and play P&P layer

Feature Based Adaptive Augmentation is presented as a plug and play (P&P) based layer; P&P implies that it can be used with any GCL method and attributed graph dataset to create one view or both. Figure 2 illustrates the P&P layer placement in the graph contrastive learning paradigm. We present the P&P layer as $T(\cdot)$, which performs augmentation on feature level with FebAA and structure level by removing edges uniformly.

**Feature level augmentation** Given that input Graph $Q = (X, A)$, feature level augmentation is only performed on nodes' feature matrix $X$ while adjacency matrix $A$ remains unchanged. It is to be stated that masking features do not change the graph's structure; in this case, $|V|$ and $|E|$ remain the same before and after augmentation. Equation 3 illustrates the feature level augmentation.

$$\overline{X}, A = T(X, A) = T(X), A \tag{3}$$

Where $\overline{X}$ is masked feature matrix.

**Structure level augmentation** Structure level augmentation changes the structure of the graph, so either one or both of $|V|$ and $|E|$ may change after the augmentation. We perform on adjacency matrix while feature matrix $X$ and node set $V$ remain unchanged. Equation 4 presents the general idea of structure level augmentation.

$$X, \overline{A} = T_A(X, A) = X, T_A(A) \tag{4}$$

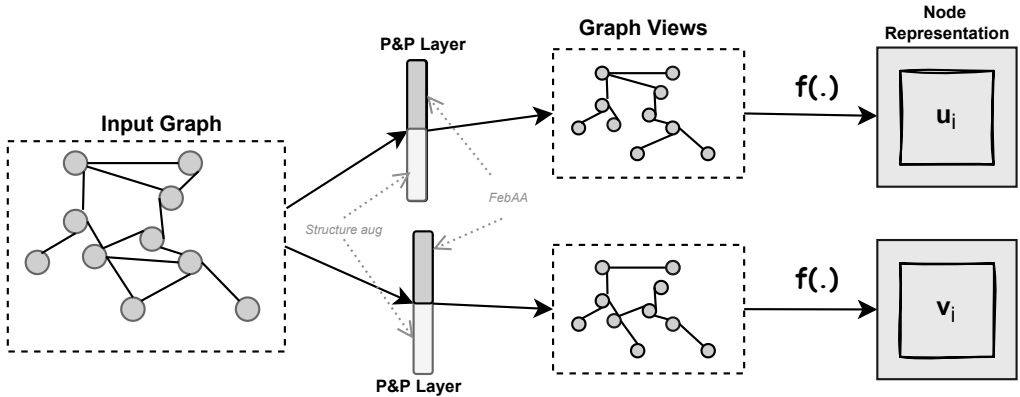

Figure 2: The placement of Plug and Play layer in GCA paradigm, Two different colors hypothetically illustrates the FebAA and structure level augmentation

Table 1: Statistics of classification benchmark datasets used in this work

| Parent | Dataset | Nodes | Features | Edges | Classes |
|---|---|---|---|---|---|
| Planetoid | Cora | 2708 | 1433 | 10556 | 7 |
| Planetoid | CiteSeer | 3327 | 3703 | 9104 | 6 |
| WikiCS | WikiCS | 11701 | 300 | 216,123 | 10 |
| Amazon | Computer | 13,752 | 767 | 491,722 | 10 |
| Amazon | Photos | 7650 | 745 | 238,162 | 8 |
| Coauthor | CS | 18,333 | 6805 | 163788 | 15 |
| Coauthor | Physics | 34,493 | 8415 | 495,924 | 5 |
| Actor | Actor | 7600 | 932 | 30019 | 5 |

where $\overline{A}$ is augmented adjacency matrix.

In this work, structure level augmentation is not adaptive, and edges are removed randomly but uniformly from the original graph. Furthermore, following the BGRL [13] and GCA [25], directed graphs are converted to the undirected graph. We use Bernoulli distribution $B(1 - p_e)$ to choose if an edge should be dropped or not. As a result, generated views have fewer edges and may become disconnected, as we do not check the connectivity after augmentation.

## 4 Experiments and results

To validate the effectiveness of our FebAA, we incorporate FebAA into GRACE [24] and BGRL [13]. We use PyGCL's [23] implementation of GRACE and BGRL's original implementation for experiments.

### 4.1 Dataset

There are eight small to medium-sized benchmark datasets used in graph representation learning research with different number of nodes, features, and edges. Properties of the graphs used in this work are mentioned in Table 1.

### 4.2 Experimental setup

Following the DGI[14], GRACE and BGRL use the linear evaluation scheme for every experiment, where each model is trained unsupervised. Embedding obtained from this step is used to train and test the $l_2$ regularized logistic regression classifier. We keep the same settings and measure accuracy on micro-averaged F1-score accuracy on transductive tasks. Furthermore, implementation details and hyper-parameter settings are presented in Appendix B. All experiments are performed using Nvidia GeForce-RTX 3090 with 24GB Ram.

Table 2: Performance measure on classification accuracy, mean and standard deviation, while X: Features, A: Adjacency matrix, S: Diffusion matrix, GRACE$_{FebAA}$ has same number of epochs as original GRACE. FebAA$_{in}$ is influential feature and FebAA$_{rn}$ is random candidate feature selection results with hyper-parameters mentioned in Appendix B.

| Dataset | Input | Cora | CiteSeer | Actor |
| --- | --- | --- | --- | --- |
| Raw features | X | 64.8 | 64.6 | N/A |
| node2vec | | 74.8 | 52.3 | N/A |
| DeepWalk | X,A | 75.7 | 50.5 | N/A |
| DeepWalk + feat. | X | 73.1 | 47.6 | N/A |
| GAE | X,A | 76.9 | 60.6 | N/A |
| VGAE | X,A | 78.9 | 61.2 | N/A |
| DGI | X,A | 82.6±0.4 | 71.8±0.7 | N/A |
| DGI | X,S | 83.8±0.5 | 72.0±0.6 | N/A |
| GRACE | X,A | 83.3±0.4 | 72.1±0.5 | 30.33±0.77 |
| GRACE$_{FebAA}$ | X,A | 86.34±0.18 | 75.0±1.72 | 30.33±0.77 |
| BGRL | X,A | 83.83±1.61 | 72.32±0.89 | 29.50 |
| MVGRL | X,S,A | 86.80±0.5 | 73.30±0.5 | N/A |
| FebAA$_{rn}$ | X,A | *85.55±1.30* | 74.09±0.85 | 30.39±0.79 |
| FebAA$_{in}$ | X,A | **87.00±0.92** | **76.26±1.46** | **30.58±1.06** |

### 4.3 Results with GRACE+FebAA

This section presents the GRACE+FebAA results with Cora, CiteSeer, and Actor datasets. We train the model for twenty runs with random data splits and present the mean and standard deviation of outputs. To generate the different contexts for the nodes in the two views, one view is generated with FebAA and the other with stochastic augmentation, while hyper-parameters are mentioned in appendix B. We use the GRACE as a baseline for this section, and Table 2 presents the results where we copy the previous results from GRACE, BGRL, and MVGRL [4]. GRACE and BGRL use feature matrix and adjacency matrix, while MVGRL also includes a diffusion matrix for training. Overall, from the table, we can see that results significantly improve when GRACE is used with our proposed FebAA. We test GRACE+FebAA with the original GRACE's hyper-parameters (represented in the table as GRACE$_{FebAA}$) and our chosen hyper-parameters (represented in the table as FebAA$_{rn}$, FebAA$_{in}$); table 2 illustrates that FebAA causes significant improvement in the Cora and CiteSeer accuracy with 5.68% and 5.24%, respectively and reasonable improvement in the Actor dataset.

### 4.4 Results with BGRL+FebAA

This section presents the BGRL+FebAA results with five node classification benchmark datasets: WikiCS, Amazon-Computers, Amazon-Photos, Coauthor-CS, and Coauthor-Physics. For this section, we use BGRL as the baseline and by following its experiments methodology, we perform twenty random data splits and model initialization; further hyper-parameters details can be found in Appendix B. Table 3 presents the results of our experiments; we see that the BGRL+FebAA combination outperforms all of the state of the art methods. There are two BGRL+FebAA flavors FebAA$_{rn}$ and FebAA$_{in}$ in the table; the farmer is based on randomly chosen candidate features and later is influential-based candidate feature. In both cases, FebAA outperforms the other methods indicating that subset feature matrix-based masking is superior to whole feature matrix-based masking.

GCA [25] is an adaptive augmentation based method and presents three types of augmentations degree centrality, Pagerank centrality and Eigenvector centrality. BGRL also deployed these three adaptive augmentation strategies, but the results were not convincing. On the contrary, our method's important achievement is that BGRL+FebAA improve five out of five datasets while stochastic augmentation based BGRL results are improved on four out of the five datasets. An important observation from Table 3 is highest results are with FebAA$_{rn}$, which selects features randomly as candidate features, cutting the pre-training cost of finding influential features. It also raises a question of, if selected features are actual important features or there is a need of another algorithm to find influential features. Also, finding influential features takes time and resources consuming; for the cost of finding the influential features, we suggest consulting section 4.5

Table 3: Results on five benchmark datasets, bold results are highest with FebAA$_{in}$ and inclined are $2^{nd}$ highest with FebAA$_{rn}$

| Dataset | WikiCS | Am.Comp | Am.Photo | Co.CS | Co.Phy |
|---|---|---|---|---|---|
| Raw features | 71.98±0 | 73.81±0 | 78.53±0 | 90.37±0 | 93.58±0 |
| DeepWalk | 74.35±0.06 | 85.68±0.06 | 89.44±0.11 | 84.61±0.22 | 91.77±0.15 |
| DeepWalk + feat. | 77.21±0.03 | 86.28±0.07 | 90.05±0.08 | 87.70±0.04 | 94.90±0.09 |
| DGI | 75.35±0.14 | 83.95±0.47 | 91.61±0.22 | 92.15±0.63 | 94.51±0.52 |
| GMI | 74.85±0.08 | 82.21±0.31 | 90.68±0.17 | OOM | OOM |
| MVGRL | 77.52±0.08 | 87.52±0.11 | 91.74±0.07 | 92.11±0.12 | 95.33±0.03 |
| GRACE | 80.14±0.48 | 89.53±0.35 | 92.78±0.45 | 91.12±0.20 | OOM |
| GCA | 78.35±0.05 | 88.94±0.15 | 92.53±0.16 | 93.10±0.01 | 95.73±0.03 |
| BGRL | 79.98±0.10 | 90.34±0.19 | 93.17±0.30 | 93.31±0.13 | 95.73±0.05 |
| FebAA$_{rn}$ | *80.32±0.40* | *90.55±0.21* | *93.43±0.23* | *93.44±0.10* | *95.81±0.07* |
| FebAA$_{in}$ | **80.59±0.58** | **91.08±0.20** | **93.77±0.20** | **93.62±0.13** | **95.89±0.09** |
| Supervised GCN | 77.19±0.12 | 86.51±0.54 | 92.42±0.22 | 93.03±0.31 | 95.65±0.16 |

### 4.5 Time complexity

FebAA has two modules to choose candidate features, random selection FebAA$_{rn}$ and as per influence FebAA$_{in}$. It is a pre-training cost as feature selection is performed before training. Hypothetically, FebAA$_{rn}$ module cost is the same as stochastic augmentation as there is negligible extra processing required to choose candidate features randomly. On the contrary, in the case of FebAA$_{in}$, it is time consuming to find the important features of any dataset depending on the size of the dataset. Below is the time complexity to find feature ranking set $R_f$; as this process is before training, it is mentioned as pre-training time $T_p$.

$$T_p = F \times i \times n \tag{5}$$

where $F$ depicts the number of features of a dataset, $i$ is the number of epochs, and $n$ is the number of runs to take an average. It is to be noticed that $T_p$ is the number of iterations not time, as time is dependent on available hardware resources. We use $i = 150$ and $n = 3$ while $F$ value is dataset dependent given in Table 1. Equation 5 is for pre-training time $T_p$ performed at feature selection from Figure 1. In contrast, masking is performed at training time so, *theoretically*, FebAA training time should be equal to stochastic augmentation's training time. It is to be notified that we provide ranked features $R_f$ of all datasets used in this research work in $CSV$ format and suggest using them for further analysis to save $T_p$. If our provided $R_f$ are used, FebAA training time will be equal to stochastic augmentation.

## 5 Candidate features selection rate analysis

Our proposed idea indicates that rather than masking from the whole feature matrix, a better approach will be to identify a subset of candidate features, and masking should be performed on candidate features. This proposal leads to another question, how many features can be in the candidate feature set? We explore the answer to this question in this section.

Three hyper-parameters affect the accuracy: feature masking ratio $F_r^s$, which decides the size of candidate features subset, feature masking probability $F_p^s$ which is the probability to mask from candidate features, and starting position $pos$ of important features. Figure 3.a to 3.h indicates that $F_r^s$ and $F_p^s$ have a definite effect on accuracy and in this section we analyze this effect. The analyzed results are from eight datasets, where all datasets are executed with 12 different combinations, summing up to 96 different variances of dataset, $F_r^s$, $F_p^s$, and $pos$.

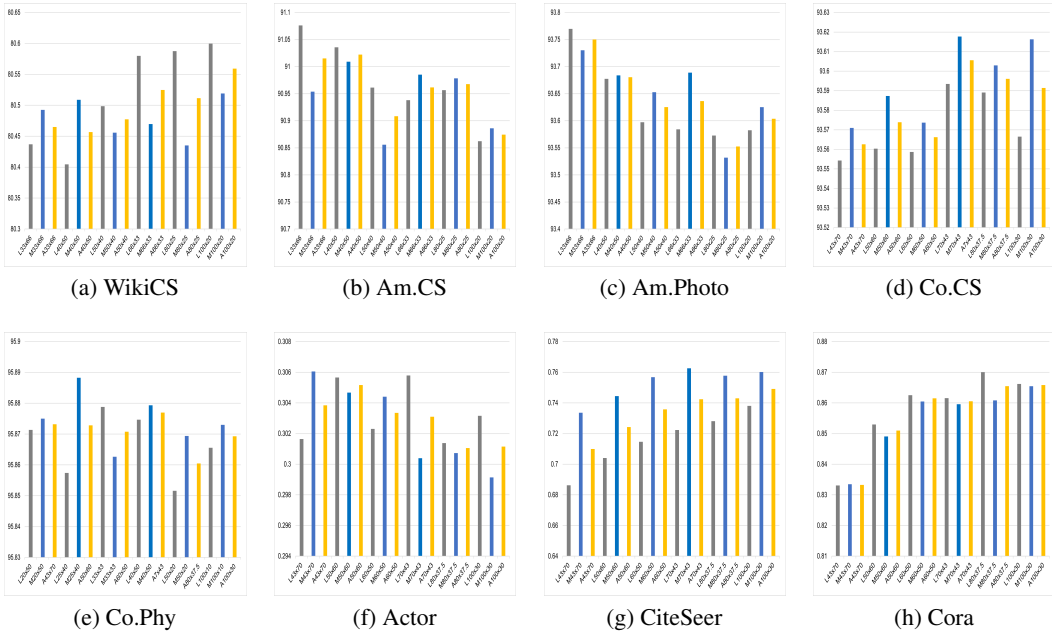

Figure 3: Relationship between feature masking ratio $F_r^s$ and results (L: Least, M: Most, A: Average, 100x20 pattern is feature masking probability $F_p^s$, and $F_r^s$ indicating mask 100% of 20% features). Horizontal axis of each sub-figure presents the percentage of features and vertical axis are accuracy.

Table 4: Feature influence masking starting position $pos$ analysis per dataset

| Dataset | $pos = l$ | $pos = m$ |
|---|---|---|
| WikiCS | 4 (66.67%) | 2 (33.3%) |
| Am.Comp | 3 (50%) | 3 (50%) |
| Am.Photo | 2 (33.3%) | 4 (66.67%) |
| Co.CS | 0 (0%) | 6 (100%) |
| Co.Phy | 1 (16.67%) | 5 (83.33%) |
| Cora | 5 (83.33%) | 2 (16.67%) |
| CiteSeer | 0 (0%) | 6 (100%) |
| Actor | 4 (66.67%) | 2 (33.3%) |
| Total | 19 (39.58%) | 29 (60.42%) |

### 5.1 Feature influence based analysis

Figure 3 represents the results of the starting position $pos$, where grey bars present $pos = l$ and blue presents $pos = m$. We compare one to one results, presenting where feature masking ratio $F_r^s$ and feature masking probability $F_p^s$ are the same but starting position $pos$ is the opposite. Overall, there are 48 experiments with both starting positions, 29 times $pos = m$ gives better accuracy than $pos = l$, 19 times; indicating if $pos = m$, accuracy will be improved oppose to $pos = l$. This analysis is not enough to state that every time setting $pos = m$ is a better choice but results illustrated in Figure 3 assure that not all features are equal and feature masking ratio $F_r^s$ affects the accuracy actively. Table 4 summarizes the overall and per dataset results in terms of starting position $pos$.

### 5.2 Feature masking ratio and probability

Feature masking ratio and feature masking probability are two inter-dependent hyper-parameters; as Figure 3 illustrates, when one's value increases other decreases. Yellow bars in the figure show the average of each combination of feature masking ratio and feature masking probability. We can observe from figure that some starting position *pos* pairs have better accuracy than others, but there are no apparent patterns under current hyper-parameter settings. Furthermore, different datasets have different features, which may affect both hyper-parameters. We believe that splitting them and analyzing them one by one can elaborate better, which we leave for future work.

This section summarize, different feature masking ratio $F_r^s$ value outputs different results. Same feature masking ratio $F_r^s$ value if applied on different features of feature matrix, it outputs different results. Although finding important features is time consuming task but our results indicates and this section concludes that *all features are not equally important* and masking a subset of feature matrix is better than masking the whole feature matrix.

## 6 Conclusion and future directions

State of the art graph contrastive learning methods remove edges and mask features to create the views, where masking is performed uniformly. We argue that uniform stochastic augmentation can corrupt the critical features leading to lesser accuracy. In this paper, we introduced the Feature Based Adaptive Augmentation (FebAA) and used it with eight small to medium scaled datasets in the transductive setting, with the state of the art methods GRACE [24] and BGRL [13] and improved the accuracy on all eight datasets. We deploy it as Plug-and-Play, as it can be used with more algorithms in the future, and we are optimistic it will improve the results. Due to limited resources, we could not test it on larger datasets; if someone has better resources is welcome to try. This work is only feature-based, and it is costly to find the influential features. We believe that the same methodology can be applied to the edge and node levels. In our following paper, we aim to improve this process. We also applied our FebAA to create both graph views, but we did not get the satisfying results; we assume that causes less diversity in both views resulting lower accuracy.

## References

[1] Jiangxia Cao, Xixun Lin, Shu Guo, Luchen Liu, Tingwen Liu, and Bin Wang. Bipartite graph embedding via mutual information maximization. *CoRR*, abs/2012.05442, 2020.

[2] Matthias Fey and Jan Eric Lenssen. Fast Graph Representation Learning with PyTorch Geometric. *arXiv e-prints*, page arXiv:1903.02428, March 2019.

[3] William L. Hamilton, Rex Ying, and Jure Leskovec. Representation learning on graphs: Methods and applications. *CoRR*, abs/1709.05584, 2017.

[4] Kaveh Hassani and Amir Hosein Khasahmadi. Contrastive multi-view representation learning on graphs. In Hal Daumé III and Aarti Singh, editors, *Proceedings of the 37th International Conference on Machine Learning*, volume 119 of *Proceedings of Machine Learning Research*, pages 4116–4126. PMLR, 13–18 Jul 2020.

[5] Weihua Hu, Bowen Liu, Joseph Gomes, Marinka Zitnik, Percy Liang, Vijay S. Pande, and Jure Leskovec. Pre-training graph neural networks. *CoRR*, abs/1905.12265, 2019.

[6] Yizhu Jiao, Yun Xiong, Jiawei Zhang, Yao Zhang, Tianqi Zhang, and Yangyong Zhu. Sub-graph contrast for scalable self-supervised graph representation learning. *CoRR*, abs/2009.10273, 2020.

[7] Baoyu Jing, Chanyoung Park, and Hanghang Tong. HDMI: high-order deep multiplex infomax. *CoRR*, abs/2102.07810, 2021.

[8] Zekarias T. Kefato and Sarunas Girdzijauskas. Self-supervised graph neural networks without explicit negative sampling. *CoRR*, abs/2103.14958, 2021.

[9] Wanyu Lin, Hao Lan, and Baochun Li. Generative causal explanations for graph neural networks. *CoRR*, abs/2104.06643, 2021.

[10] Kaili Ma, Haochen Yang, Han Yang, Tatiana Jin, Pengfei Chen, Yongqiang Chen, Barakeel Fanseu Kamhoua, and James Cheng. Improving graph representation learning by contrastive regularization. *CoRR*, abs/2101.11525, 2021.

[11] Adam Paszke, Sam Gross, Francisco Massa, Adam Lerer, James Bradbury, Gregory Chanan, Trevor Killeen, Zeming Lin, Natalia Gimelshein, Luca Antiga, Alban Desmaison, Andreas Kopf, Edward Yang, Zachary DeVito, Martin Raison, Alykhan Tejani, Sasank Chilamkurthy, Benoit Steiner, Lu Fang, Junjie Bai, and Soumith Chintala. Pytorch: An imperative style, high-performance deep learning library. In H. Wallach, H. Larochelle, A. Beygelzimer, F. d'Alché-Buc, E. Fox, and R. Garnett, editors, *Advances in Neural Information Processing Systems*, volume 32. Curran Associates, Inc., 2019.

[12] Jiezhong Qiu, Qibin Chen, Yuxiao Dong, Jing Zhang, Hongxia Yang, Ming Ding, Kuansan Wang, and Jie Tang. GCC: graph contrastive coding for graph neural network pre-training. *CoRR*, abs/2006.09963, 2020.

[13] Shantanu Thakoor, Corentin Tallec, Mohammad Gheshlaghi Azar, Mehdi Azabou, Eva L. Dyer, Rémi Munos, Petar Veličković, and Michal Valko. Large-scale representation learning on graphs via bootstrapping, 2021.

[14] Petar Veličković, William Fedus, William L. Hamilton, Pietro Liò, Yoshua Bengio, and Devon Hjelm. Deep graph infomax. In *ICLR 2019*, May 2019.

[15] Pengfei Wei, Zhenzhou Lu, and Jingwen Song. Variable importance analysis: A comprehensive review. *Reliability Engineering & System Safety*, 142:399–432, 2015.

[16] Lirong Wu, Haitao Lin, Zhangyang Gao, Cheng Tan, and Stan Z. Li. Self-supervised on graphs: Contrastive, generative, or predictive. *CoRR*, abs/2105.07342, 2021.

[17] Zonghan Wu, Shirui Pan, Fengwen Chen, Guodong Long, Chengqi Zhang, and Philip S. Yu. A comprehensive survey on graph neural networks. *IEEE Transactions on Neural Networks and Learning Systems*, 32(1):4–24, 2021.

[18] Yaochen Xie, Zhao Xu, Zhengyang Wang, and Shuiwang Ji. Self-supervised learning of graph neural networks: A unified review. *CoRR*, abs/2102.10757, 2021.

[19] Yuning You, Tianlong Chen, Yongduo Sui, Ting Chen, Zhangyang Wang, and Yang Shen. Graph contrastive learning with augmentations. In H. Larochelle, M. Ranzato, R. Hadsell, M. F. Balcan, and H. Lin, editors, *Advances in Neural Information Processing Systems*, volume 33, pages 5812–5823. Curran Associates, Inc., 2020.

[20] Jiaqi Zeng and Pengtao Xie. Contrastive self-supervised learning for graph classification. *CoRR*, abs/2009.05923, 2020.

[21] Hanlin Zhang, Shuai Lin, Weiyang Liu, Pan Zhou, Jian Tang, Xiaodan Liang, and Eric P. Xing. Iterative graph self-distillation. *CoRR*, abs/2010.12609, 2020.

[22] Qi Zhu, Yidan Xu, Haonan Wang, Chao Zhang, Jiawei Han, and Carl Yang. Transfer learning of graph neural networks with ego-graph information maximization. *CoRR*, abs/2009.05204, 2020.

[23] Yanqiao Zhu, Yichen Xu, Qiang Liu, and Shu Wu. An empirical study of graph contrastive learning. *CoRR*, abs/2109.01116, 2021.

[24] Yanqiao Zhu, Yichen Xu, Feng Yu, Qiang Liu, Shu Wu, and Liang Wang. Deep graph contrastive representation learning. *CoRR*, abs/2006.04131, 2020.

[25] Yanqiao Zhu, Yichen Xu, Feng Yu, Qiang Liu, Shu Wu, and Liang Wang. Graph contrastive learning with adaptive augmentation. *CoRR*, abs/2010.14945, 2020.

## A  Ablation Studies

Table 5 present the ablation study on the FebAA, first column has dataset used in this study and second presents the original results. Third column of the table presents FebAA results, where edges are dropped stochastic manner and features are dropped with our proposed adaptive augmentation. Last column of table is OFD stands for only feature dropping and no edges are dropped. We observe and as previous research demonstrates, when hybrid (features and edges) augmentation is used, graph contrastive learning algorithms performance is improved.

Table 5: Ablation Study, *OFD*= Only features drop and *Original* are original paper (BGRL and GRACE) results

| Dataset | Original | FebAA | FebAA(OFD) |
|---|---|---|---|
| WikiCS | 79.98±0.10 | 80.59±0.58 | 80.40±0.51 |
| Am.Comp | 90.34±0.19 | 91.08±0.20 | 90.86±0.24 |
| Am.Photo | 93.17±0.30 | 93.77±0.20 | 93.31±0.18 |
| Co.CS | 93.31±0.13 | 93.62±0.13 | 93.61±0.12 |
| Co.Phy | 95.73±0.05 | 95.89±0.09 | 95.84±0.10 |
| Cora | 83.3±0.4 | 87.00±0.92 | 86.86±1.02 |
| CiteSeer | 72.1±0.5 | 76.26±1.46 | 75.97±1.30 |
| Actor | - | 30.58±1.06 | 30.74±1.09 |

Table 6: Hype-parameters for FebAA$_{in}$, row 1 is for view 1 and row 2 for view 2 against each dataset, while $F_d \times F_r$ represent drop probability and drop ratio respectively

| Dataset | $F_d \times F_r$ | Least | Feature drop | Edge drop | Training epochs | Learning rate | Weight decay | Hidden layers |
|---|---|---|---|---|---|---|---|---|
| WikiCS | 0.8×0.25<br>- | M<br>- | 0.2<br>0.1 | 0.2<br>0.3 | 10,000 | $5 \times 10^{-4}$ | $1^{-5}$ | 512 |
| Am.Comp | 0.3×0.66<br>- | L<br>- | 0.2<br>0.1 | 0.5<br>0.4 | 10,000 | $5 \times 10^{-4}$ | $1^{-5}$ | 512 |
| Am.Photo | 0.3×0.66<br>- | L<br>- | 0.1<br>0.2 | 0.4<br>0.1 | 10,000 | $1 \times 10^{-5}$ | $1^{-5}$ | 512 |
| Co.CS | 0.7×0.43<br>- | M<br>- | 0.3<br>0.4 | 0.3<br>0.2 | 10,000 | $1 \times 10^{-5}$ | $1^{-5}$ | 512 |
| Co.Phy | 0.25×0.4<br>- | M<br>- | 0.1<br>0.4 | 0.4<br>0.1 | 10,000 | $1 \times 10^{-5}$ | $1^{-5}$ | 512 |
| Cora | -<br>0.80×0.375 | -<br>L | 0.4<br>0.3 | 0.4<br>0.2 | 2000 | $5 \times 10^{-3}$ | $10^{-5}$ | 128 |
| CiteSeer | -<br>0.7×0.43 | -<br>M | 0.4<br>0.3 | 0.4<br>0.2 | 2000 | $1 \times 10^{-3}$ | $10^{-5}$ | 256 |
| Actor | -<br>0.7×0.43 | -<br>L | 0.3<br>0.3 | 0.3<br>0.3 | 2000 | $5 \times 10^{-4}$ | $10^{-5}$ | 128 |

## B Implementation

**Implementation and Datasets:** FebAA is implemented using PyTorch Geometric 11.1 [2], PyTorch [11] and PyGCL [23]. Datasets used in this study are accessible via PyTorch Geometric libraries.

**Hyper-parameters:** Dataset used in this study are small to medium sized dataset and for training purposes different hyper-parameters are used. For datasets used in BGRL training epochs are 10,000 and for Grace's dataset 2000 epochs, while further details are summarized in Table 6. Hyper-parameter configuration files, FebAA codes, important features of each dataset, and random seeds to regenerate results; are available at https://github.com/mhadnanali/FebAA.